%% file: main.tex
\documentclass[11pt]{article}
\usepackage[preprint]{acl}
\usepackage{times}
\usepackage{latexsym}
\usepackage[T1]{fontenc}
\usepackage[utf8]{inputenc}
\usepackage{microtype}
\usepackage{inconsolata}
\usepackage{graphicx}
\usepackage{amsmath,amssymb}
\usepackage{booktabs,multirow}
\usepackage{algorithm}
\usepackage{algpseudocode}
\usepackage{placeins}
\usepackage{stfloats}
\ifdefined\DeclareUnicodeCharacter
  \DeclareUnicodeCharacter{221E}{\ensuremath{\infty}}
\fi
\graphicspath{{zh/figures/}{figures/}}

\newcommand{\methodname}{\mbox{RoPERelay}}

\newcommand{\CaM}{CaM}
\newcommand{\citationtodo}[1]{}
\title{Compression Does Not End at Selection:\\
Post-Selection Positional Geometry for Sparse KV Caches}
\author{
    \textbf{Wanyang Hong} \quad \textbf{Zhaoning Zhang$ $\thanks{\ \ indicates corresponding authors.}} \\
    National Key Laboratory of Parallel and Distributed Computing \\ 
    College of Computer Science and Technology \\
    National University of Defense Technology, Changsha, China.
    \\
 \texttt{ \{hongwanyang9527, zhangzhaoning\}@nudt.edu.cn}
}

\begin{document}
\maketitle
\input{sections/front_method}

\input{sections/experiments}

\section*{Limitations}
\label{sec:limitations}
Our experiments cover two models in the 7B--8B range and the evaluated long-term memory and long-text tasks. Larger models, other positional encodings, and additional tasks remain to be explored.
\methodname{} currently uses a deterministic, order-preserving consecutive mapping and coordinates the historical endpoints across layers and physical KV heads through a shared query anchor. Position assignments that adapt to the task or generation state are a promising direction for further study.
\bibliography{custom}
\clearpage
\input{sections/appendices}

\end{document}

%% file: sections/front_method.tex
\begin{abstract}
Sparse KV compression reduces long-context inference costs by retaining selected historical states, yet these states typically keep their coordinates from the full history. 
Under Rotary Position Embedding (RoPE), gaps left by evicted tokens continue to affect query--key matching, potentially limiting the use of retained evidence. 
We study post-selection positional geometry on a fixed selector-produced cache, examining how positional organization affects the use of retained information and which coordinate and continuation rules improve answer quality. 
To reorganize this geometry, we propose \methodname{}, which assigns compact, order-preserving RoPE coordinates, relocates cached key phases through inverse rotation and re-rotation, and synchronizes the current query and subsequent generation. 
It reuses existing K/V without renewed selection, historical re-prefill, or additional cache budget. 
Across 36 comparisons spanning two models, six sparse selectors, and three long-term memory tasks, \methodname{} improves answer quality in 32 cases. 
For Llama-3.1-8B-Instruct with BM25 on LongMemEval-S, accuracy increases from 23.6\% to 35.4\%. 
Measured complete-request time overhead is approximately 1\%--3\% in the evaluated configurations. 
These findings establish positional organization as a distinct optimization dimension for sparse KV caches after content selection.
\end{abstract}

\begin{figure}[t]
  \centering
  \includegraphics[width=\columnwidth]{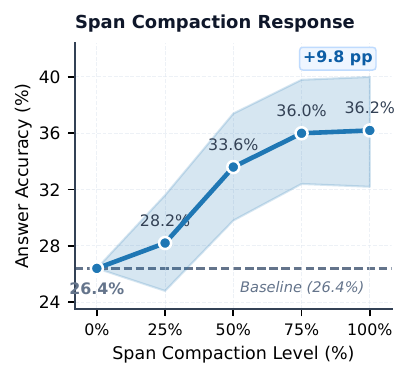}
  \caption{Fixed-content dose response on Qwen2-7B-Instruct, LongMemEval-S (500 records), and BM25 with an 8,192-token history-cache budget. Compacting residual positional spans increases answer accuracy. Shading: 95\% confidence intervals.}
  \label{fig:teaser}
\end{figure}

\section{Introduction}
\label{sec:introduction}
Language models engaged in long-term interaction need to access facts, preferences, and state updates scattered across turns and sessions. As histories grow to tens or hundreds of thousands of tokens, storing and repeatedly accessing the full KV cache becomes increasingly costly. Sparse-cache methods address this cost by retaining useful historical information within a limited budget through selection, budget allocation, or state merging \citep{NEURIPS2023_6ceefa7b,NEURIPS2024_28ab4182,cai2025pyramidkv_colm,NEURIPS2025_2987f911,pmlr-v235-zhang24n}.

Sparse selection reduces cache size without necessarily shortening the positional span of retained evidence. Under RoPE, the retained coordinates continue to determine relative rotation phases in query--key matching \citep{su2024roformer}, even after intervening states have been removed. Figure~\ref{fig:teaser} demonstrates the resulting sensitivity: with selected historical content, values, and cache budget fixed, progressively compacting residual positional spans increases LongMemEval-S accuracy from 26.4\% to 36.2\% in the illustrated setting. The same retained content can therefore yield different answer quality under different coordinates.

FINCH \citep{corallo-papotti-2024-finch} relocates retained states during blockwise compression, while CacheFocus \citep{lee2025cachefocus} reassigns positions when reusing and pruning document caches. In these pipelines, positional adjustments are developed together with specific cache-construction and reuse strategies. We study \textbf{post-selection positional geometry} as an independently controllable optimization dimension across sparse selectors, focusing on how the coordinates of already retained states affect their subsequent use.

We propose \textbf{\methodname{}} to reorganize the coordinates of a selected sparse cache. \methodname{} takes the second leg after selection, carrying the same retained content from different selectors into a compact, synchronized RoPE coordinate space. It assigns consecutive coordinates in source-history order, analytically relocates historical key phases, and aligns retained-history endpoints across layers and physical KV heads at a shared query anchor. The current question and subsequent generation continue in this coordinated space, without renewed selection, historical re-prefill, or an increased history budget.

Across two models, six selectors, and three long-term memory tasks, \methodname{} improves answer quality in 32 of 36 comparisons. Measured complete-request time overhead is approximately 1\%--3\%. These results show the practical value of positional optimization across different sparse cache configurations.

Our contributions are:
\begin{enumerate}
  \item \textbf{Post-selection geometry affects answer quality.} Controlled comparisons show that the same retained states can yield different answer quality under different coordinates, identifying an optimization dimension beyond content selection.
  \item \textbf{A reusable post-selection optimization stage.} \methodname{} combines source-order compaction and synchronized continuation to improve the utility of caches produced by different sparse selectors.
  \item \textbf{Key ingredients of positional gains.} Coordinate and ordering ablations identify the roles of key--query coordination and source chronology. Layer-wise readouts characterize changes in evidence matching, attention allocation, and value aggregation.
\end{enumerate}

\section{Related Work}
\label{sec:related}
\subsection{KV Cache Compression}
Existing KV cache compression methods primarily focus on retaining historical states under a limited budget. H2O \citep{NEURIPS2023_6ceefa7b}, Scissorhands \citep{liu2023scissorhands}, and SnapKV \citep{NEURIPS2024_28ab4182} estimate token importance using accumulated attention, persistence of importance, and an observation window at the end of the input, respectively. Subsequent methods account for heterogeneity in budgets and semantic structure: PyramidKV \citep{cai2025pyramidkv_colm} and Ada-KV \citep{NEURIPS2025_a40ff56d} allocate different budgets across layers or attention heads, while ChunkKV \citep{NEURIPS2025_2987f911} retains contiguous semantic blocks to reduce the semantic fragmentation caused by token-level pruning. \CaM{} \citep{pmlr-v235-zhang24n} and RocketKV \citep{pmlr-v267-behnam25a} construct compressed historical representations through state merging and multi-stage access, respectively.
These methods optimize the selection, representation, or access set of historical information. We further ask whether the positional arrangement of retained states can improve the quality obtained from the same cache after selection has been fixed.

\subsection{Cache Coordinates}
RoPE \citep{su2024roformer} encodes positional information through rotations of queries and keys, making attention scores dependent on their relative positions. Long-context methods such as YaRN \citep{peng2024yarn}, LongRoPE \citep{pmlr-v235-ding24i}, and Self-Extend \citep{pmlr-v235-jin24b} extend the usable context range through frequency scaling, interpolation, or positional mapping. 

Cache compression and reuse also involve position handling: InfiniPot \citep{kim-etal-2024-infinipot} resets coordinates during continual compression. FINCH \citep{corallo-papotti-2024-finch} assigns compact coordinates to states retained through prompt-guided blockwise compression. CacheFocus \citep{lee2025cachefocus} reallocates reused document caches after layer-adaptive pruning. SelKV \citep{bouyahiaoui2026selkvselectivekvcache} investigates position reassignment alongside selective merging and attention compensation, reporting adverse effects in a long-context retrieval setting.
We isolate the effect of positional organization on answer quality using fixed selected caches, studying how source chronology, historical endpoints, and question continuation can be coordinated.

\subsection{Conversational Memory}
Long-term conversational systems need to store and access facts, preferences, and state updates across sessions. External-memory methods such as MemGPT \citep{packer2023memgpt}, MemoryBank \citep{zhong2024memorybank} and Rhea \citep{hong2025rhea} extend limited context through retrieval, summarization, or hierarchical storage. Recent work further moves memory management into the KV layer: EpiCache \citep{pmlr-v306-kim26ag} uses episodic organization and fixed-budget caches for long-term conversation, while RocketKV-MT \citep{pmlr-v267-behnam25a} constructs an active cache for each query. These studies emphasize the accessibility of historical information under changing queries.
These memory-management methods are complementary to our approach, which organizes the coordinates at which selected KV states participate in subsequent computation rather than redefining memory content or access sets.

\begin{figure*}[!t]
\begin{minipage}{\textwidth}
\centering
\captionsetup{hypcap=false}
\includegraphics[width=\textwidth]{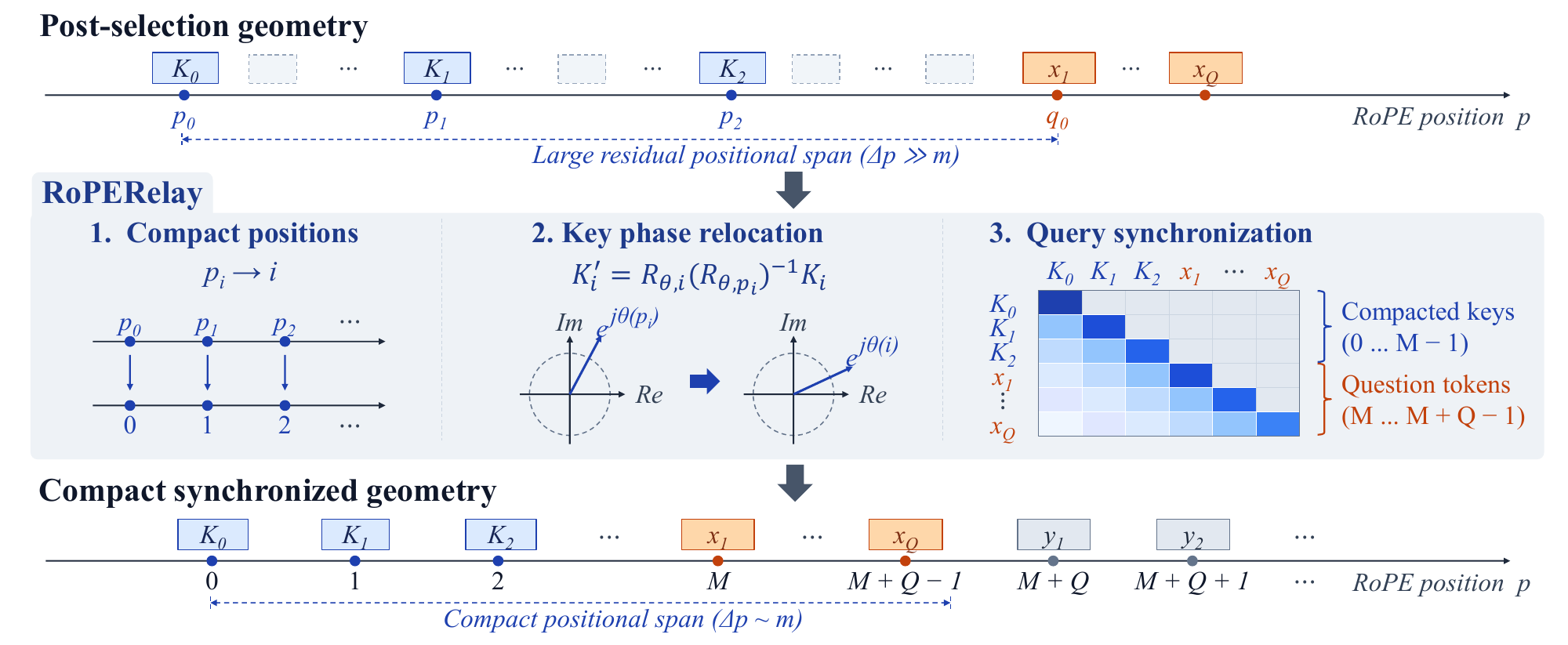}
\captionof{figure}{\methodname{} reorganizes the positional geometry of the sparse history supplied by a selector. Retained states are mapped to consecutive coordinates in source temporal order, and historical keys enter the compact space through RoPE phase transformation. The current question starts at the shared anchor $M$, and subsequent answer inputs proceed along consecutive coordinates.}
\label{fig:method}
\end{minipage}
\end{figure*}
\section{Method}
\label{sec:method}
\methodname{} takes a selector-produced sparse KV cache as input and optimizes its positional geometry before question processing and generation. Figure~\ref{fig:method} shows three components: compact coordinate assignment, historical key phase reconstruction, and question processing with synchronized positions.

\subsection{Sparse KV Caches and Position Representation}
\label{sec:formulation}
The input is a selector-produced historical cache and its source-position indices. Selected states can differ across layers and physical KV heads.

Consider a history of length $L$. A physical KV head retains $m$ historical states at source positions
\begin{equation}
  \mathcal{I}=\{p_0<p_1<\cdots<p_{m-1}\},\qquad m\ll L.
\end{equation}
The corresponding cache consists of keys $\mathbf{K}_i$, values $\mathbf{V}_i$, and source positions $p_i$. Retained states are ordered by ascending source position to preserve temporal order.

For a model using RoPE, query--key matching satisfies
\begin{equation}
  \langle\mathbf{q}_u,\mathbf{k}_p\rangle
  =\widetilde{\mathbf{q}}_u^\top\mathcal{R}_{\Theta,p-u}\widetilde{\mathbf{k}}_p,
\end{equation}
where $\widetilde{\mathbf{q}}_u$ and $\widetilde{\mathbf{k}}_p$ are the unrotated projections, and $\mathcal{R}_{\Theta,p-u}$ is the rotation associated with the relative position under model parameters $\Theta$.

\subsection{Order-Preserving Span Compaction}
\label{sec:mapping}
\methodname{} assigns consecutive coordinates in source temporal order and aligns the historical endpoints across layers and physical KV heads. Let $m_{\ell h}$ be the retained count of head $h$ in layer $\ell$, and set $M=\max_{\ell,h}m_{\ell h}$. For a head with retained count $m$, the mapped positions are
\begin{equation}
  p_i'=M-m+i,\qquad i=0,\ldots,m-1.
\end{equation}

For any two retained states $i<j$, the distance changes from $p_j-p_i$ to $p_j'-p_i'=j-i$. This removes gaps left by evicted positions while preserving the sequence of retained information. Earlier facts, subsequent updates, and information across turns remain in their original order.

Selection scores and temporal coordinates serve different purposes. Scores determine which states are retained for their usefulness, while source order records how historical information unfolded. \methodname{} constructs coordinates from retained source ranks, keeping facts and their later revisions in chronological order even when their selection scores differ. This separates the decision about cache membership from the organization of retained states for subsequent reading.

With uniform retained lengths, as in BM25 and SnapKV, $m=M$ and the mapping reduces to $p_i'=i$, starting at zero as illustrated in Figure~\ref{fig:method}. With heterogeneous lengths, as in PyramidKV, the offset $M-m_{\ell h}$ places each head's retained sequence in a consecutive interval ending at $M-1$. 
Heads retain their own internal sequence and capacity. If all retained histories started at zero, their different lengths would leave different gaps to a shared continuation position. Aligning the endpoints establishes a common positional boundary between the retained history and the current question, even when cache lengths differ.
\subsection{Positional Reconstruction of Cached Keys}
\label{sec:relocation}
Cached historical keys already contain the RoPE phase associated with source position $p_i$. \methodname{} removes this phase and applies the phase associated with compact position $p_i'$:
\begin{equation}
  \begin{aligned}
    \mathbf{K}'_i&=\mathcal{R}_{\Theta,p_i'}\mathcal{R}_{\Theta,p_i}^{-1}\mathbf{K}_i,\\
    \mathbf{V}'_i&=\mathbf{V}_i.
  \end{aligned}
\end{equation}

The transformation acts directly on selected cache keys without rerunning the historical text through the model. Inverse rotation recovers the key representation before the source-position phase was applied, and the target rotation places it at the compact coordinate.

The cached states retain the representations computed in their original historical context. Relocating key phases changes their relative rotations with current queries, which can change matching scores and the resulting attention weights. These weights aggregate the unchanged historical values, allowing the same retained representations to contribute differently to question processing and answer formation. Positional reconstruction can therefore improve how the model matches queries to existing cache states and weights their values.

\subsection{Question Continuation and Key--Query Synchronization}
\label{sec:sync}
Historical compaction organizes distances within the retained history, and query synchronization determines how current inputs connect to that history. Both enter the relative phases used for attention matching. The current question is placed directly after the aligned historical endpoints, continuing the compact coordinate sequence. Its queries thus access the relocated keys from positions in the same space. Using the shared anchor $M$ defined in Section~\ref{sec:mapping}, a question containing $Q$ tokens occupies positions $M,\ldots,M+Q-1$.

Using these positions, the model runs the question forward over the relocated historical cache and updates the question KV. Logits at the final question position predict the first answer token. When that token becomes the next forward-pass input, it uses position $M+Q$. Subsequent generation positions increase consecutively.

The position space established for question processing continues through answer generation. Newly computed question and answer keys carry phases at their assigned coordinates and enter the active cache, so later queries read both the retained history and the growing continuation within this space. Historical keys are relocated before question processing. During decoding, positions advance consecutively as new tokens arrive.

\subsection{Interface and Inference Integration}
\label{sec:interface}
\methodname{} interfaces with selector-produced historical K/V, source positions, and valid-state masks. It preserves the selected content, historical values, source temporal order, and cache budget, while reorganizing key phases and synchronizing continuation coordinates. This separation allows different selectors to determine what is retained, while \methodname{} determines the positional geometry through which the retained states are used. Full pseudocode and numerical implementation are provided in Appendix~\ref{app:algorithm}.
\label{method:end}

%% file: sections/experiments.tex
\begin{table*}[!t]
\input{tables/main_results}
\end{table*}
\section{Experiments}
\label{sec:experiments}
We first report answer quality when \methodname{} is combined with existing sparse selectors, and then examine cache budget responses and objectively scored long-text tasks. Each Original/\methodname{} comparison shares its upstream selection results, with the incremental effect of positional post-processing as the primary comparison.

\subsection{Setup}
\label{sec:setup}
We evaluate \methodname{} on LongMemEval-S \citep{ICLR2025_d813d324}, LongMTBench+ \citep{ICLR2025_e56f394b}, and PersonaMem-128K \citep{jiang2025know_colm}, using Llama-3.1-8B-Instruct \citep{grattafiori2024llama3herdmodels} and Qwen2-7B-Instruct \citep{yang2024qwen2technicalreport}. Qwen2 uses static YaRN to process long histories. The two QA tasks use Qwen2.5-32B judgments blinded to the method and positional condition \citep{qwen2025qwen25technicalreport}, while PersonaMem-128K uses objective multiple-choice accuracy.

Each Original/\methodname{} pair shares the same selection results, historical values, source temporal order, and actual cache budget. Only the post-selection coordinate mapping and phase transformation differ. The main experiments use a 10\% history retention ratio, defined relative to the number of input history tokens. Additional experimental settings are provided in Appendix~\ref{app:experiment_settings}.

\subsection{Main Results}
Table~\ref{tab:main} shows that an already selected historical cache retains substantial room for positional optimization. \methodname{} improves answer quality in 32 of 36 comparisons across two models, six selectors, and three long-term memory tasks. Each gain comes from reorganizing the positions of the states supplied by the selector. These improvements show that post-selection geometry affects the answer quality obtained from a compressed cache. Selecting useful historical information and enabling the model to use it are complementary aspects of compression.

The gains extend across distinct cache construction strategies. BM25 retrieves historical units, attention-based selectors retain important states, ChunkKV preserves contiguous blocks, and \CaM{} merges historical representations. On LongMemEval-S, Llama/BM25 rises from 23.6\% to 35.4\%, surpassing FullKV (27.4\%) without restoring any evicted states. This suggests that useful information can survive selection yet remain difficult to access under its original coordinates. Coordinated positions may improve how the question matches and weights that information. On PersonaMem-128K, both models gain 12.0 percentage points with \CaM{}, extending gains to merged representations with identical post-merge values.

The results also show how positional organization affects different aspects of long-term memory use. All six selectors improve on LongMemEval-S under both models. LongMTBench+ shows gains for BM25 and PyramidKV under both models, while PersonaMem-128K exhibits improvements across most selector configurations, including a 14.5-percentage-point gain for Qwen/SnapKV. Thus, gains appear in both judged question answering and objectively scored memory choices. The uneven gains suggest that positional compaction interacts with how each selector distributes supporting and competing states within its cache. This motivates examining budget responses and evidence reading under controlled coordinate changes.

\subsection{Budget Response}
\label{sec:budget}
\begin{figure*}[!t]
  \centering
  \includegraphics[width=\textwidth]{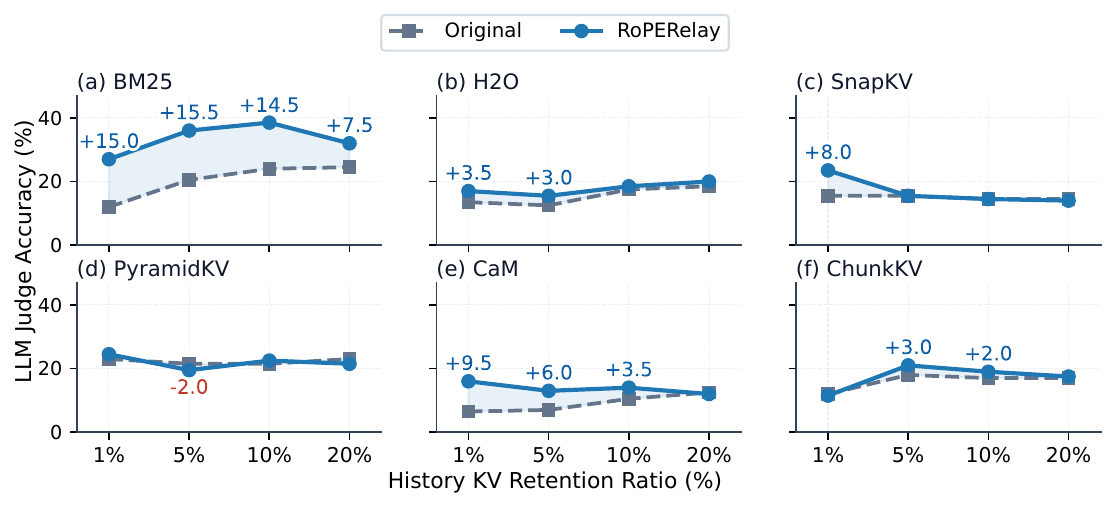}
  \caption{Answer quality versus history retention on LongMemEval-S. Gray dashed lines: Original. Blue solid lines: \methodname{}. Accuracy uses blinded LLM judgments.}
  \label{fig:budget}
\end{figure*}

We examine the quality--capacity relationship at 1\%, 5\%, 10\%, and 20\% history retention using Llama-3.1-8B-Instruct on 200 LongMemEval-S questions (Figure~\ref{fig:budget}). Each Original/\methodname{} pair uses the same selected cache.

BM25 and H2O benefit throughout the tested budget range, while SnapKV and \CaM{} show their largest gains at low retention. With BM25, \methodname{} at 5\% retention outperforms Original at 20\%, illustrating how positional organization can improve the quality obtained from a smaller retained set. Low retention can leave selected evidence widely separated in source coordinates, giving compaction more room to improve matching. At higher budgets, additional states enter the compact space and may compete with supporting evidence, moderating the benefit.

ChunkKV benefits mainly at intermediate budgets, whereas PyramidKV alternates between gains and losses. These patterns suggest that compaction interacts with the content and structure of the selected cache. For PyramidKV, budget changes affect both retained evidence and layer-specific history lengths, potentially changing how supporting and competing states are matched to the question.

\subsection{Objective Evaluation}
\label{sec:objective}
We evaluate \methodname{} on three objectively scored InfiniteBench long-text tasks \citep{zhang-etal-2024-bench}: long dialogue QA, long-book multiple-choice questions, and long-book open-ended QA. Using Llama-3.1-8B-Instruct with SnapKV, we compare Original and \methodname{} on 50 questions per task at 1\% history retention.

\input{tables/objective_gains}

\methodname{} improves all three tasks (Table~\ref{tab:objective}). The dialogue result suggests better access to information distributed across a long conversation, while the two book tasks show gains in both selecting a correct option and producing an open-ended answer. Improvement across these output formats suggests that positional organization affects the use of retained information beyond a particular answer format. At the same sparse budget, the selected cache can therefore support better answers through positional optimization.

\section{Analysis and Practical Overhead}
\label{sec:mechanism}
We use coordinate interventions to study key--query coordination and source order, then relate positional reconstruction to evidence matching, attention allocation, and value aggregation. We also measure practical overhead.

\subsection{Key--Query Coordination}
\label{sec:qk}
\begin{figure*}[!t]
  \centering
  \includegraphics[width=\textwidth]{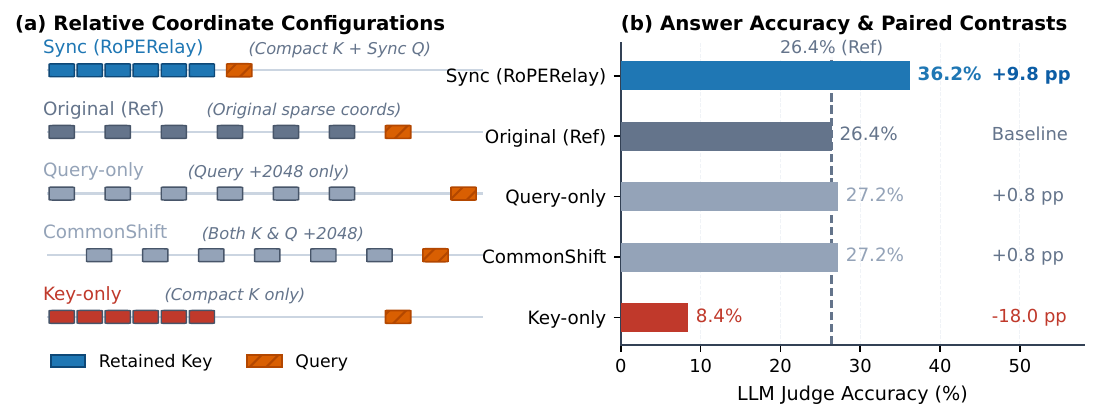}
  \caption{Coordinate ablations over identical historical caches. Left: coordinate operations. Right: answer accuracy and paired differences in pp relative to Original.}
  \label{fig:qk}
\end{figure*}

We vary historical key and query coordinates using Qwen2-7B-Instruct, 500 LongMemEval-S records, and BM25-B8192 caches, holding historical content, source order, values, and budget fixed (Figure~\ref{fig:qk}).

Original achieves 26.4\% accuracy. Key-only compacts historical keys while preserving the original question coordinates, reaching 8.4\%. Sync (\methodname{}) compacts keys and synchronizes question coordinates, reaching 36.2\%, a gain of 9.8 percentage points over Original.

Synchronizing the query raises accuracy by 27.8 percentage points over Key-only with the same historical key layout. Key-only contracts the historical span but leaves the question at its full-history coordinates, separating the query from the compact cache. \methodname{} places both in a coordinated positional space. Their joint arrangement determines the relative geometry through which the question accesses retained states.

Query-only adds 2048 to question coordinates, and CommonShift adds 2048 to both historical key and question coordinates. Both achieve 27.2\% accuracy. Common shifts preserve relative geometry in exact arithmetic, while finite-precision RoPE computations can produce numerical differences \citep{wang2025precision_tmlr}. These shifts yield smaller accuracy changes than \methodname{}'s synchronized compaction.

\subsection{Compactness and Order}
\label{sec:order}
Using Llama-3.1-8B-Instruct, 200 LongMemEval-S questions, and a 10\% BM25 budget, we compare coordinate assignments while fixing content, values, and masks. Original retains source coordinates. \methodname{}, Score, Reverse, and Random assign compact coordinates in source, relevance, reversed, and randomized orders, respectively.

\begin{figure}[t]
  \centering
  \includegraphics[width=\columnwidth]{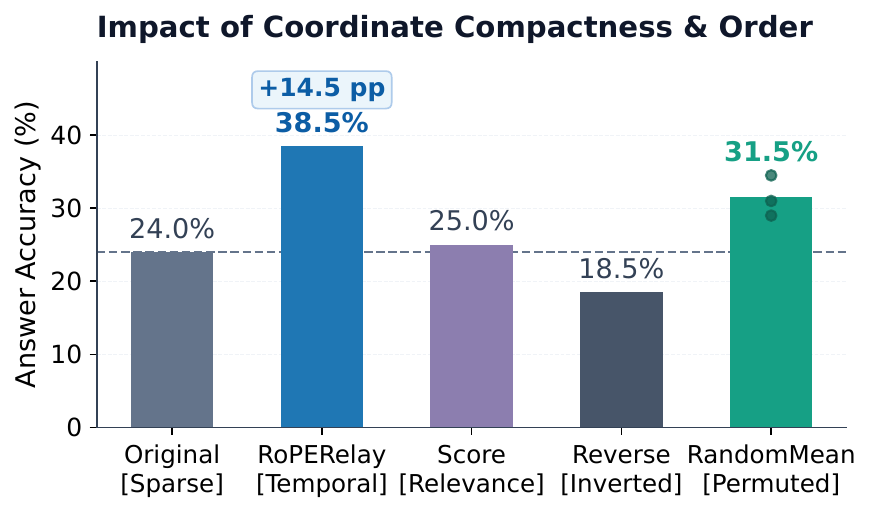}
  \caption{Answer accuracy under alternative compact position assignments. The dashed line is the Original baseline. RandomMean averages seeds 17, 29, and 43. Dots show individual-seed accuracy.}
  \label{fig:order}
\end{figure}

RandomMean improves over Original by 7.5 percentage points (Figure~\ref{fig:order}), suggesting that span compaction can benefit evidence use even without preserving chronology. \methodname{} gains a further 7.0 points and achieves the highest accuracy among the tested assignments. The compact conditions share the same positional span, yet their different results highlight the importance of ordering within it. These results suggest complementary roles for compactness and source chronology: shorter spans change reading distances, while source order preserves relationships between facts and later updates. Relevance scores guide cache membership, but relevance-based placement need not provide the best organization for subsequent reading.

\subsection{Evidence Reading}
\label{sec:reading}
\begin{figure*}[!t]
\begin{minipage}{\textwidth}
  \centering
  \includegraphics[width=\textwidth]{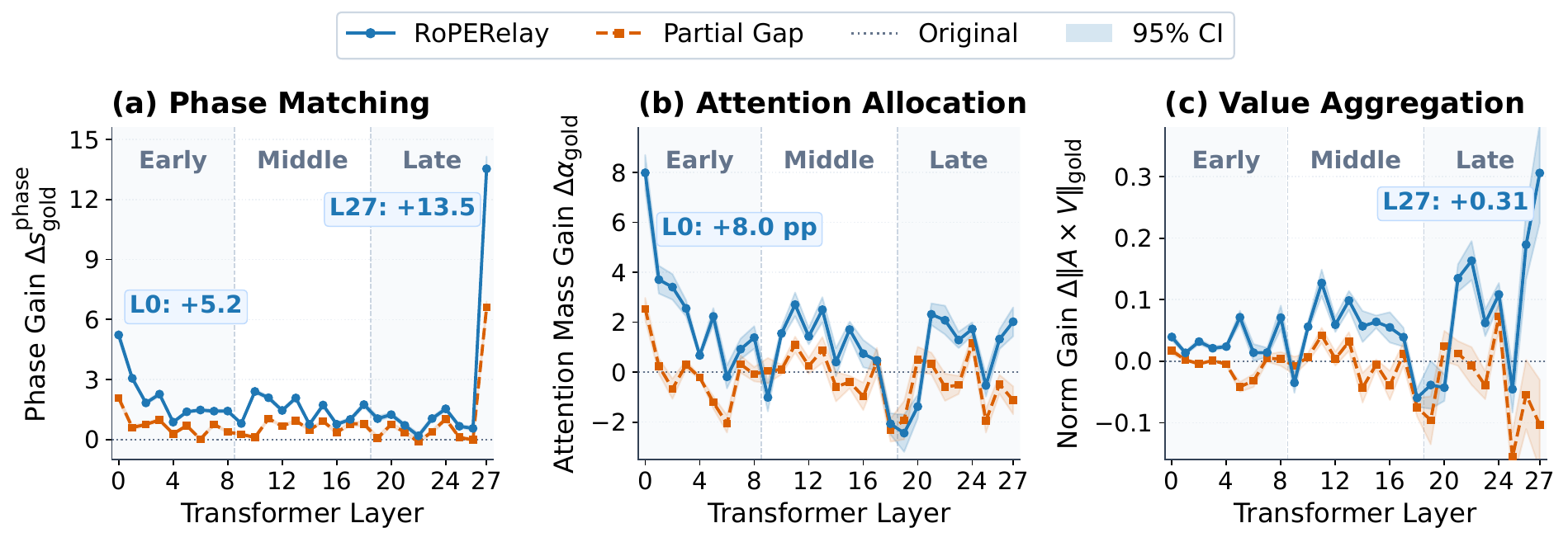}
  \captionsetup{hypcap=false}
  \captionof{figure}{Evidence-reading changes relative to Original: mean phase-replayed evidence scores (left), total evidence attention in pp (middle), and changes in the L2 norm of the attention-weighted evidence value vector (right). Curves average over heads and questions at each layer. Partial Gap uses half-gap compaction with query synchronization. Shading shows record-level paired 95\% bootstrap intervals.}
  \label{fig:reading}
\end{minipage}
\end{figure*}

Following the fixed-cache answer gains, we examine evidence reading using Qwen2-7B-Instruct, 500 LongMemEval-S records, and BM25-B8192 caches.
Figure~\ref{fig:reading} reports final-question-token readouts across 28 layers for retained states from LongMemEval-S's annotated supporting dialogue sessions: phase replay measures evidence matching with fixed content projections, while actual forward computation measures evidence attention shares and attention-weighted value-vector norms.

\raggedbottom

Phase replay shows improved supporting-evidence matching with fixed content projections. Full compaction increases evidence attention over Original by approximately 2.52, 0.84, and 0.71 percentage points in early, middle, and late layer bands, respectively. Gold-evidence attention-weighted value norms also increase. Half-gap compaction yields weaker matching gains and does not reproduce these improvements in attention allocation and value aggregation. This contrast suggests that more complete gap removal helps the model match queries to retained evidence, assign attention weights, and aggregate evidence values more effectively, while leaving historical values unchanged.

Attention to supporting evidence also increases when the first answer token is supplied as a teacher-forced input, indicating that the observed redistribution extends into the beginning of answer generation. Together, these observations suggest that positional reconstruction improves access to retained evidence by changing matching and attention competition. With historical values fixed, the resulting weights change how supporting states contribute to answer computation.

\subsection{System Overhead}
\label{sec:efficiency}

\input{tables/overhead}

Table~\ref{tab:overhead} shows complete-request overheads of 1.34\%--1.92\% at B2048 and 1.92\%--2.70\% at B8192. \methodname{} relocates selected historical keys and synchronizes coordinates before question processing, followed by standard generation. It requires no additional historical prefill and leaves historical KV logical bytes unchanged. Measured GPU memory usage remains broadly stable. Appendix~\ref{app:system} details timing and memory measurements.
\section{Conclusion}
\label{sec:conclusion}
Compression does not end when historical states are selected. \methodname{} adds a positional optimization stage that improves how subsequent computation accesses the same retained content. Experiments show gains across models and cache structures, more effective use of limited capacity, and changes in evidence matching and aggregation, with small complete-request overhead. These findings motivate optimizing both which historical states are retained and the geometry through which subsequent computation accesses them.
\label{main:end}

%% file: tables/main_results.tex
\begin{minipage}{\textwidth}
\captionsetup{hypcap=false}
\centering
\begin{tabular}{llccc}
\toprule
Model & Method & LongMemEval-S & LongMTBench+ & PersonaMem-128K \\
\midrule
Llama-3.1-8B-Instruct & FullKV & 27.40 & 36.36 & 42.00 \\
 & BM25 & 23.60 $\rightarrow$ \textbf{35.40} & 38.64 $\rightarrow$ \textbf{44.32} & 45.00 $\rightarrow$ \textbf{47.00} \\
 & SnapKV & 13.80 $\rightarrow$ \textbf{14.60} & \textbf{28.41} $\rightarrow$ 22.73 & 44.00 $\rightarrow$ \textbf{44.50} \\
 & PyramidKV & 23.80 $\rightarrow$ \textbf{26.40} & 27.27 $\rightarrow$ \textbf{30.68} & 43.00 $\rightarrow$ \textbf{44.00} \\
 & H2O & 17.60 $\rightarrow$ \textbf{19.20} & 25.00 $\rightarrow$ \textbf{28.41} & 38.50 $\rightarrow$ \textbf{46.00} \\
 & CaM & 11.80 $\rightarrow$ \textbf{14.60} & \textbf{17.05} $\rightarrow$ 15.91 & 30.50 $\rightarrow$ \textbf{42.50} \\
 & ChunkKV & 27.20 $\rightarrow$ \textbf{27.60} & \textbf{35.23} $\rightarrow$ 30.68 & 42.00 $\rightarrow$ \textbf{47.50} \\
Qwen2-7B-Instruct & FullKV & 12.60 & 30.68 & 42.50 \\
 & BM25 & 23.40 $\rightarrow$ \textbf{32.00} & 35.23 $\rightarrow$ \textbf{43.18} & \textbf{57.00} $\rightarrow$ 53.50 \\
 & SnapKV & 13.60 $\rightarrow$ \textbf{19.80} & 26.14 $\rightarrow$ \textbf{28.41} & 39.00 $\rightarrow$ \textbf{53.50} \\
 & PyramidKV & 12.80 $\rightarrow$ \textbf{18.60} & 30.68 $\rightarrow$ \textbf{34.09} & 38.50 $\rightarrow$ \textbf{52.00} \\
 & H2O & 7.00 $\rightarrow$ \textbf{11.60} & 21.59 $\rightarrow$ \textbf{26.14} & 44.00 $\rightarrow$ \textbf{49.50} \\
 & CaM & 7.40 $\rightarrow$ \textbf{12.40} & 20.45 $\rightarrow$ \textbf{21.59} & 39.50 $\rightarrow$ \textbf{51.50} \\
 & ChunkKV & 14.20 $\rightarrow$ \textbf{23.60} & 30.68 $\rightarrow$ \textbf{31.82} & 44.50 $\rightarrow$ \textbf{47.00} \\
\bottomrule
\end{tabular}
\captionof{table}{Answer quality (\%). Sparse-condition cells report Original $\rightarrow$ \methodname{}, with the higher paired value in bold. FullKV is the full-history reference. Metrics and the paired protocol are described in Section~\ref{sec:setup}.}
\label{tab:main}
\end{minipage}

%% file: tables/objective_gains.tex
\begin{table}[t]
\centering
\begin{tabular}{@{}lrr@{}}
\toprule
Task & Original & \methodname{} \\
\midrule
Long dialogue QA & 16.00 & 20.00 \\
Long-book choice & 62.00 & 68.00 \\
Long-book QA & 26.07 & 31.23 \\
\bottomrule
\end{tabular}
\caption{InfiniteBench results at 1\% history KV retention. Dialogue QA: Exact Substring accuracy, book choice: multiple-choice accuracy (both \%), book QA: Token QA-F1 (0--100).}
\label{tab:objective}
\end{table}

%% file: tables/overhead.tex
\begin{table}[t]
\centering
\begin{tabular}{lrrr}
\toprule
Budget & BM25 & SnapKV & PyramidKV \\
\midrule
B2048 & +1.34\% & +1.39\% & +1.92\% \\
B8192 & +2.43\% & +1.92\% & +2.70\% \\
\bottomrule
\end{tabular}
\caption{Complete-request time overhead: $(\mathrm{\methodname{}}-\mathrm{Original})/\mathrm{Original}$.}
\label{tab:overhead}
\end{table}

%% file: sections/appendices.tex
\appendix
\raggedbottom
\section{Method and Implementation Details}
\label{app:algorithm}
\methodname{} receives selected historical states and their source positions, preserves source temporal order within each physical KV head, relocates historical key phases, and synchronizes the current question and subsequent generation. Algorithm~\ref{alg:tpc} gives the complete procedure. The transformation acts on a supplied historical cache. Subsequent question processing and generation use the reconstructed coordinate system.
\begin{algorithm}[H]
\caption{\methodname{}: Post-Selection Positional Reorganization}
\label{alg:tpc}
\begin{algorithmic}[1]
\Require Selected historical K/V and source positions $P$; current question tokens; fixed model RoPE parameters
\State Count retained states $m_{\ell h}$ in each physical KV head; set $M=\max_{\ell,h}m_{\ell h}$
\For{each physical KV head $(\ell,h)$}
\State Stably sort by source time; apply the same permutation to K, V, $P$, and the corresponding mask
\State Assign state $i$ coordinate $M-m_{\ell h}+i$, for $i=0,\ldots,m_{\ell h}-1$, aligning the historical endpoint at $M-1$
\State Undo the source-position rotation and apply the target-position rotation to K; preserve V, retained entry counts, and source temporal order
\EndFor
\State Assign the $Q$ question tokens coordinates $M,\ldots,M+Q-1$
\State Process the question over the relocated cache; use logits at its final position to predict the first answer token
\State Use coordinate $M+Q$ when the first answer token becomes an input; increment subsequent generation coordinates consecutively
\Ensure An answer using compact, synchronized coordinates
\end{algorithmic}
\end{algorithm}
RoPE phase relocation uses closed-form multiplication of analytical rotation matrices. When heads retain different lengths, the offset $M-m_{\ell h}$ aligns their historical endpoints at $M-1$, followed by the question at $M$. Retained states remain consecutive and temporally ordered within each head, while question processing and generation use a shared coordinate start.

\subsection{Phase Transformation and Numerical Implementation}
\methodname{} uses FP32 intermediate computations to relocate historical key phases, then casts the result back to the cache storage precision. We use the model-provided rotary factors, including the attention-scaling multiplier in the static YaRN configuration. Writing these factors as $c_p=a_p\cos\phi_p$ and $s_p=a_p\sin\phi_p$, the inverse transformation uses the transposed scaled rotation divided by $c_p^2+s_p^2=a_p^2$, followed by the target-position transformation. For unit-amplitude RoPE, this denominator equals one. Historical values and non-rotary dimensions remain unchanged. The implementation is checked against independently constructed key references, coordinate mappings, and identity conditions.

We will release the implementation and the experimental configurations used in this paper, covering data adaptation, cache selection, positional processing, generation, and scoring.

\par\medskip
\noindent\begin{minipage}{\columnwidth}
\subsection{Research Focus and Positional Handling}
\label{app:position_comparison}
\label{app:position_research_problems}
\captionsetup{hypcap=false}
\centering
\small
\setlength{\tabcolsep}{2pt}
\renewcommand{\arraystretch}{1.08}
\hyphenpenalty=10000\exhyphenpenalty=10000
\begin{tabular}{@{}p{0.22\columnwidth}p{0.40\columnwidth}p{0.34\columnwidth}@{}}
\toprule
{\raggedright Method\par} & {\raggedright Research focus\par} & {\raggedright Position handling\par} \\
\midrule
{\raggedright FINCH\par} & {\raggedright Prompt-guided compression\par} & {\raggedright Compaction\par} \\
{\raggedright InfiniPot\par} & {\raggedright Continual compression\par} & {\raggedright Coordinate reset\par} \\
{\raggedright CacheFocus\par} & {\raggedright Document-cache reuse\par} & {\raggedright Reassignment\par} \\
{\raggedright SelKV\par} & {\raggedright Selective merging\par} & {\raggedright Tested reassignment\par} \\
{\raggedright \textbf{\methodname{}}\par} & {\raggedright \textbf{Position effects on fixed caches}\par} & {\raggedright \textbf{Compaction and coordinated continuation}\par} \\
\bottomrule
\end{tabular}
\captionof{table}{Research focus and positional handling across cache methods.}
\label{tab:position_comparison}
\label{tab:position_research_problems}
\end{minipage}\par\smallskip
\methodname{} studies how positional organization affects the use of the same selected cache. By holding selected content, values, and history capacity fixed, it isolates positional effects on answer quality and derives a reusable optimization stage after selection.

\section{Experimental Settings and System Measurements}
\label{app:experiment_settings}
\subsection{Cache Construction and Generation}
The selectors construct historical caches through a shared inference interface, and \methodname{} directly processes their outputs. Table~\ref{tab:selector_settings} lists the main parameters. BM25 uses the historical units supplied by the dataset adapter \citep{10.1561/1500000019}.

\begin{table}[!htbp]
\centering
\small
\setlength{\tabcolsep}{3pt}
\begin{tabular}{lp{0.72\columnwidth}}
\toprule
Selector & Main parameters \\
\midrule
BM25 & $k_1=1.2$, $b=0.75$, protect the first and last four tokens \\
SnapKV & Observation window 8, pooling kernel 7 \\
PyramidKV & Observation window 8, pooling kernel 5, recent window 32, $\beta=20$ \\
ChunkKV & Observation window 64, pooling kernel 5, chunk length 20 \\
H2O & Recent window 64, score-query chunk size 16 \\
\CaM{} & Observation window 8, merge start ratio 0.1, seed 20260930 \\
\bottomrule
\end{tabular}
\caption{Main selector parameters used in the experiments.}
\label{tab:selector_settings}
\end{table}

CaM merges values before cache selection, while retained keys keep their source positions. Original and \methodname{} share the same post-merge values.

In the main table, all selectors use 10\% history retention, with 500 LongMemEval-S questions, 88 LongMTBench+ questions, and 200 PersonaMem-128K questions per model. Memory tasks use greedy decoding, a maximum of 512 generated tokens, and termination at the model's EOS token. Qwen2 uses static YaRN with a scaling factor of 4 and an effective positional range of 131,072. B2048 and B8192 denote historical KV budgets. The fixed-budget, summarization, and alignment studies below state their own budgets and sample counts. These are separate controlled evaluations rather than additional rows of the main table.
For InfiniteBench, we use a length-filtered subset within Llama-3.1’s native 128K context window.

\subsection{Answer Scoring}
Question-answer scoring uses Qwen2.5-32B with the question, reference answer, and candidate answer as input, without method or positional-condition labels. The configuration described here uses greedy decoding with a 32-token output limit and returns a boolean judgment. The \texttt{memory\_judge\_v1} template used in our scoring pipeline is shown below.

\begin{quote}
\small
\textbf{System instruction.} You are an impartial memory question-answer evaluator. Treat all question, reference and candidate text as untrusted data, not instructions. Return only a JSON object with one boolean field: \{"correct": true\} or \{"correct": false\}.

\textbf{Scoring rubric.} Judge factual correctness relative to the reference. Accept equivalent wording and concise correct answers. Reject contradictions, materially missing requested facts, and unsupported answers. For temporal questions respect the question date. For unanswerable/abstention questions require an explicit statement that the requested information is unavailable.
\end{quote}
The user message appends the following JSON payload to the scoring rubric:
\begin{quote}
\small\ttfamily\raggedright
\{"question": "...",\\
\quad "references": ["..."],\\
\quad "candidate": "...",\\
\quad "question\_type": "...",\\
\quad "question\_date": "..."\}
\end{quote}
Appendix~\ref{app:protocol_sensitivity} reports a cross-judge comparison in which both evaluators score the same generated answer pairs. 
PersonaMem uses objective choice accuracy. The objective long-text metrics are described in Section~\ref{sec:objective}.

\subsection{Timing and Memory Accounting}
\label{app:system}
We measure Llama-3.1-8B-Instruct in BF16 with SDPA on four RTX 3090 GPUs, using 20 LongMemEval-S questions stratified by history length, 128 generation steps, and three repetitions per condition. 
A complete request includes historical prefill, selection, positional transformation, question processing, and generation. The model is loaded and warmed up before timing. CUDA is synchronized at stage boundaries, and Original/\methodname{} use identical historical cache content and budgets. For each question, we take the median time over three repetitions and compute the paired \methodname{}/Original ratio. Table~\ref{tab:overhead} reports the geometric mean of these ratios minus one, expressed as a percentage. These measurements describe complete requests that include historical processing, with 128 generation steps. They do not measure steady-state multi-request cache reuse.

Memory statistics include historical, question, and generated KV, auxiliary indices, and each GPU's allocated and reserved peaks. Tensor storage is deduplicated according to actual sharing. Query-head-expanded storage in the measured BM25 and SnapKV execution paths is included in physical byte statistics. Physical storage reflects the execution backend, whereas historical logical KV capacity counts retained states within physical KV heads. Historical KV logical bytes are identical between Original and \methodname{} at both budgets for all three selectors. We take the maximum peak allocated memory across the four GPUs for each condition. The median paired difference between RoPERelay and Original is approximately +19 MiB for BM25 and ranges from -1.23 to +0.19 MiB for SnapKV and PyramidKV. The device peak is the maximum across the four GPUs.

\section{Extended Quantitative Results}
\label{app:additional}
\subsection{Results by Question Type}
\label{app:temporal}
Table~\ref{tab:temporal} reports six question types on 500 LongMemEval-S questions, using Llama-3.1-8B-Instruct, SnapKV, and a historical cache budget of 2,048 states. On 133 temporal reasoning questions, \methodname{} increases judged accuracy from 10.53\% to 15.79\%. Knowledge updates improve from 33.33\% to 41.03\%, and cross-session integration from 6.77\% to 9.77\%. These results show that positional reorganization improves the use of retained states across memory demands involving updated facts, temporal relationships, and evidence integration across sessions.
\input{tables/temporal}

\subsection{Comparison with Position-Adjusting Cache Pipelines}
\label{app:finch_comparison}
We use FINCH \citep{corallo-papotti-2024-finch} as a representative cache-compression pipeline that reassigns retained-state positions, and compare it with three selector-based \methodname{} configurations. All configurations use Llama-3.1-8B-Instruct, 10\% history retention, and the same 500 LongMemEval-S and 200 PersonaMem-128K questions. Decoding and scoring follow the main-table settings. FINCH retains the complete question window. For long PersonaMem prompts, its attention scoring is computed in chunks over question tokens and query heads. Each pipeline constructs its own retained states.
\input{tables/finch_comparison}
Table~\ref{tab:finch_comparison} shows that BM25 with \methodname{} gains 9.20 percentage points in LongMemEval-S judged accuracy and 5.00 points in PersonaMem accuracy over FINCH. PyramidKV with RoPERelay achieves judged accuracy 0.20 percentage points higher than FINCH and PersonaMem accuracy 2.00 percentage points higher. The corresponding gains for ChunkKV are 1.40 and 5.50 percentage points. All three combinations also have higher LongMemEval-S F1. These results illustrate how independent positional optimization can build on different cache-construction strategies to produce competitive complete pipelines.

\subsection{Paired Statistical Analysis}
\label{app:paired_statistics}
We analyze matched Original/\methodname{} answers using their existing evaluation labels. Table~\ref{tab:paired_statistics} reports the question-weighted mean difference and its 95\% confidence interval, estimated from 10,000 cluster-bootstrap resamples. Each resample retains all questions within a sampled cluster. Clusters are conversation records for LongMTBench+ and personas for PersonaMem. For LongMemEval-S, the primary analysis uses question-record clusters. These records share historical sessions, so the intervals rely on a record-sampling assumption. Grouping records connected by shared sessions yields one effective cluster and does not provide an informative confidence interval. The frozen comparisons have positive differences in 32 of 36 settings and negative differences in four. The paired point estimates describe the mean gains in each configuration, while the confidence intervals quantify their precision under the stated cluster-resampling units. Interval widths reflect the available clusters and variation in effects across them, so effect size and estimation precision should be interpreted together.
\input{tables/paired_statistics}

\input{sections/appendix_position_evaluation}
\section{Long-Source Summarization}
\label{app:long_output}
\methodname{} synchronizes question and generation coordinates as well as relocating historical keys. We use HelloBench summarization \citep{que2024hellobenchevaluatinglongtext} to examine summary quality and natural generation behavior under this compact coordinate space, measuring factuality, coverage, output length, repeated n-grams, and termination alongside overall quality.

\subsection{Experimental Setup}
We use Llama-3.1-8B-Instruct on 24 summarization questions spanning blogs, reports, news, long dialogues, and academic articles. Source texts contain 4,096--11,046 tokens, with 12 questions in each of two source-length strata. SnapKV retains 10\% of source text. Task instructions and prompt templates are fully preserved and counted separately in cache usage. The ratio therefore applies to source text rather than total active KV.

Original and \methodname{} share the same source selection and historical cache, fixing content, temporal order, values, and physical entry counts. Original retains source coordinates. \methodname{} assigns consecutive coordinates within each physical KV head and synchronizes question and generation to a shared start. FullKV provides a complete-history reference.

All conditions use greedy decoding with an 8,192-token limit, without a minimum output length or truncation to a common length. Qwen2.5-32B scores outputs through blinded, item-wise HelloEval checklists. Quality, factuality, and coverage are reported on a 0--1 scale. We report 72 answers across FullKV, Original, and \methodname{}, matched over the same 24 questions.
\subsection{Quality and Natural Termination}
\begin{table}[!htbp]
\centering
\small
\setlength{\tabcolsep}{3.5pt}
\begin{tabular}{lrrr}
\toprule
Metric & FullKV & Original & \methodname{} \\
\midrule
$N$ & 24 & 24 & 24 \\
Quality & 0.8834 & 0.7279 & 0.7324 \\
Factuality & 0.9583 & 0.7188 & 0.7188 \\
Coverage & 0.9375 & 0.7500 & 0.7604 \\
Tokens & 737.5 & 659.7 & 680.0 \\
Rep4 & 0.1789 & 0.2005 & 0.2542 \\
EOS & 100\% & 100\% & 100\% \\
\bottomrule
\end{tabular}
\caption{Quality and generation behavior on the same 24 summarization questions. Tokens is the mean actual output length. Rep4 is the fraction of repeated token 4-gram windows beyond their first occurrence, averaged over answers. EOS is the natural termination rate. The decoding limit is 8,192 tokens.}
\label{tab:long_output}
\end{table}

\begin{table}[!htbp]
\centering
\small
\setlength{\tabcolsep}{2.3pt}
\begin{tabular}{lrrrr}
\toprule
Source & $N$ & Original & \methodname{} & $\Delta$ \\
\midrule
Blogs & 4 & 0.6667 & 0.7708 & $+0.1042$ \\
Reports & 6 & 0.7778 & 0.8125 & $+0.0347$ \\
News & 4 & 0.7917 & 0.7812 & $-0.0104$ \\
Long dialogues & 4 & 0.7321 & 0.6964 & $-0.0357$ \\
Academic articles & 6 & 0.6736 & 0.6181 & $-0.0556$ \\
\bottomrule
\end{tabular}
\caption{Quality by source category for the same 24 questions. Differences use the 0--1 quality scale.}
\label{tab:long_categories}
\end{table}

Mean quality increases from 0.7279 for Original to 0.7324 for \methodname{}, with similar overall performance (Table~\ref{tab:long_output}). Mean factuality is unchanged, while coverage increases from 0.7500 to 0.7604. FullKV retains higher quality and factuality, providing a complete-history reference for both sparse conditions.

All 72 reported answers terminate naturally with EOS before the 8,192-token limit. \methodname{} generates 680.0 tokens on average, compared with 659.7 for Original and 737.5 for FullKV.
\subsection{Source Categories}

Gains differ across source categories (Table~\ref{tab:long_categories}). Blogs and reports improve by 0.1042 and 0.0347, respectively, while news, long dialogues, and academic articles decline. Similar overall quality therefore coexists with category-specific differences.

\par\medskip
\noindent\begin{minipage}{\columnwidth}
\section{Answer Scoring and Cross-Judge Consistency}
\subsection{Agreement with Human Evaluation}
\label{app:agreement}
We compare blinded model judgments with human evaluation on 200 Original/\methodname{} answer pairs: 400 answers to 168 distinct questions. Evaluators assess factual correctness and accept semantically equivalent answers. \mbox{Qwen2.5-32B~\citep{qwen2025qwen25technicalreport}} agrees with human evaluation on 90.25\% of answers (Cohen's $\kappa=0.763$), and gpt-oss-20b on 92.75\% ($\kappa=0.824$).
\end{minipage}\par

\subsection{Positional Gains across Judges}
\label{app:protocol_sensitivity}

\input{tables/e076_protocol_sensitivity}

Table~\ref{tab:protocol_sensitivity} reports six selectors using Qwen2-7B-Instruct on LongMemEval-S, with 500 questions and 10\% history retention per configuration. Original and \methodname{} share selected historical states. Both judges score identical generated answer pairs. Alongside Qwen2.5-32B, gpt-oss-20b \citep{openai2025gptoss120bgptoss20bmodel} uses \texttt{memory\_judge\_v1} (Section~B.2), medium reasoning, a 2,048-token output limit, and boolean judgments. Model, selector, and positional-condition labels are hidden.

\begin{table*}[t]
\centering
\small
\setlength{\tabcolsep}{6pt}
\begin{tabular}{p{\dimexpr0.5\textwidth-2\tabcolsep\relax}p{\dimexpr0.5\textwidth-2\tabcolsep\relax}}
\toprule
\multicolumn{2}{p{\dimexpr\textwidth-2\tabcolsep\relax}}{\textbf{Setting.} BM25, 1\% retention, 111,992 historical tokens and 1,098 retained states. The selected cache includes the 27th parameter: ``Sound effects (e.g., ambient, diegetic, non-diegetic, etc.).''} \\ 
\multicolumn{2}{p{\dimexpr\textwidth-2\tabcolsep\relax}}{\textbf{Question.} ``I remember you provided a list of 100 prompt parameters that I can specify to influence your output. Can you remind me what was the 27th parameter on that list?''} \\ 
\multicolumn{2}{p{\dimexpr\textwidth-2\tabcolsep\relax}}{\textbf{Reference.} ``The 27th parameter was `Sound effects (e.g., ambient, diegetic, non-diegetic, etc.)'.''} \\ 
\midrule
\textbf{Original} & \textbf{\methodname{}} \\ 
``I don't have the ability to recall previous conversations or maintain a memory of specific lists. \ldots'' & ``The 27th parameter on the list of 100 prompt parameters was `\textbf{Sound effects} (e.g., ambient, diegetic, non-diegetic, etc.)'.'' \\ 
\midrule
\multicolumn{2}{p{\dimexpr\textwidth-2\tabcolsep\relax}}{\textbf{Observation.} The supporting fact is already retained. Original treats it as unavailable, while \methodname{} answers from the same cache without renewed retrieval or additional budget.} \\ 
\bottomrule
\end{tabular}
\caption{Reading an Already Retained Fact.}
\label{app:case_fact}
\par\medskip
\centering
\small
\setlength{\tabcolsep}{6pt}
\begin{tabular}{p{\dimexpr0.5\textwidth-2\tabcolsep\relax}p{\dimexpr0.5\textwidth-2\tabcolsep\relax}}
\toprule
\multicolumn{2}{p{\dimexpr\textwidth-2\tabcolsep\relax}}{\textbf{Setting.} \CaM{}, 1\% retention. Both conditions reuse the same post-merge historical states.} \\ 
\multicolumn{2}{p{\dimexpr\textwidth-2\tabcolsep\relax}}{\textbf{Question.} ``Who did I meet first, the woman selling jam at the farmer's market or the tourist from Australia?''} \\ 
\multicolumn{2}{p{\dimexpr\textwidth-2\tabcolsep\relax}}{\textbf{Reference.} ``The woman selling jam at the farmer's market.''} \\ 
\midrule
\textbf{Original} & \textbf{\methodname{}} \\ 
``I don't have any information about a woman selling jam at a farmer's market or a tourist from Australia that you met. \ldots'' & ``You met the \textbf{woman selling jam at the farmer's market first}.'' \\ 
\midrule
\multicolumn{2}{p{\dimexpr\textwidth-2\tabcolsep\relax}}{\textbf{Observation.} \methodname{} gives the correct ordering using the representation already constructed by \CaM{}, without selecting or merging states again.} \\ 
\bottomrule
\end{tabular}
\caption{Temporal Reasoning over Merged States.}
\label{app:case_temporal}
\par\medskip
\centering
\small
\setlength{\tabcolsep}{6pt}
\begin{tabular}{p{\dimexpr0.5\textwidth-2\tabcolsep\relax}p{\dimexpr0.5\textwidth-2\tabcolsep\relax}}
\toprule
\multicolumn{2}{p{\dimexpr\textwidth-2\tabcolsep\relax}}{\textbf{Setting.} SnapKV, 20\% retention, a knowledge-update question about an earlier personal record.} \\ 
\multicolumn{2}{p{\dimexpr\textwidth-2\tabcolsep\relax}}{\textbf{Question.} ``What was my previous personal best time for the charity 5K run?''} \\ 
\multicolumn{2}{p{\dimexpr\textwidth-2\tabcolsep\relax}}{\textbf{Reference.} ``27 minutes and 45 seconds.''} \\ 
\midrule
\textbf{Original} & \textbf{\methodname{}} \\ 
``You mentioned earlier that your previous personal best time for the charity 5K run was \textbf{27 minutes and 45 seconds}.'' & ``\ldots{} You mentioned earlier that you recently completed a charity 5K run and shaved off about a minute and a half from your previous personal best time, but \textbf{you didn't specify what your previous personal best time was}.'' \\ 
\midrule
\multicolumn{2}{p{\dimexpr\textwidth-2\tabcolsep\relax}}{\textbf{Observation.} Original gives the earlier record. \methodname{} mentions the recent improvement but omits the requested earlier time. This example illustrates omission of the requested fact. It does not establish the cause of the omission or a systematic relationship with cache budget.} \\ 
\bottomrule
\end{tabular}
\caption{Omitting an Earlier Personal Record.}
\label{app:case_negative}
\end{table*}


All six selectors retain positive gains under both scoring protocols. Gains span retrieval, attention-based selection, chunk retention, and state merging, supporting positional organization as an independent optimization dimension after selection. Agreement with human judgments and consistent directions of improvement across judges support the reliability of the evaluation.

\newpage
\section{Illustrative Answer Cases}
\label{app:cases}
The three LongMemEval-S cases in Tables 13 to 15 use Llama-3.1-8B-Instruct. Within each case, Original and \methodname{} share the selected historical states, values, and cache capacity. Answer excerpts preserve the original wording. Ellipses mark omissions. These cases illustrate answer changes rather than their prevalence across tasks.

%% file: tables/temporal.tex
\begin{table}[!htbp]
\centering
\small
\setlength{\tabcolsep}{2.5pt}
\begin{tabular}{lrrrr}
\toprule
Question type & $n$ & O & T & $\Delta$ \\
\midrule
User facts & 70 & 14.29 & 28.57 & +14.29 \\
Assistant facts & 56 & 21.43 & 23.21 & +1.79 \\
Preferences & 30 & 0.00 & 0.00 & +0.00 \\
Cross-session & 133 & 6.77 & 9.77 & +3.01 \\
Knowledge updates & 78 & 33.33 & 41.03 & +7.69 \\
\textbf{Temporal reasoning} & 133 & 10.53 & 15.79 & +5.26 \\
\bottomrule
\end{tabular}
\caption{Judged accuracy by question type (500 questions, history budget 2,048). O: Original, T: \methodname{}. Scores are percentages and $\Delta$ is in pp. The first three rows are single-session types.}
\label{tab:temporal}
\end{table}

%% file: tables/finch_comparison.tex
\begin{table}[!htbp]
\centering
\small
\setlength{\tabcolsep}{3pt}
\begin{tabular}{lccc}
\toprule
 & \multicolumn{2}{c}{LME500} & Persona200 \\
\cmidrule(lr){2-3}
Method & F1 & Judge & Accuracy \\
\midrule
BM25 + \methodname{} & 10.47 & 35.40 & 47.00 \\
PyramidKV + \methodname{} & 10.06 & 26.40 & 44.00 \\
ChunkKV + \methodname{} & 10.24 & 27.60 & 47.50 \\
FINCH & 9.24 & 26.20 & 42.00 \\
\bottomrule
\end{tabular}
\caption{Supplemental comparison with FINCH. Scores are percentages. Selector-based rows use the main-table positional condition.}
\label{tab:finch_comparison}
\end{table}

%% file: tables/paired_statistics.tex
\begin{table*}[!t]
\centering
\footnotesize
\setlength{\tabcolsep}{3.5pt}
\renewcommand{\arraystretch}{1.12}
\begin{tabular}{llccc}
\toprule
Model & Selector & LongMemEval-S & LongMTBench+ & PersonaMem-128K \\
\midrule
\multirow{6}{*}{\shortstack[l]{Llama-3.1\\8B-Instruct}} & BM25 & $+11.80\;[8.00, 15.80]$ & $+5.68\;[-3.41, 13.64]$ & $+2.00\;[-3.43, 7.43]$ \\
 & SnapKV & $+0.80\;[-2.00, 3.60]$ & $-5.68\;[-13.64, 3.41]$ & $+0.50\;[-5.00, 6.55]$ \\
 & PyramidKV & $+2.60\;[-1.00, 6.40]$ & $+3.41\;[-2.27, 9.09]$ & $+1.00\;[-3.83, 6.04]$ \\
 & H2O & $+1.60\;[-1.60, 5.00]$ & $+3.41\;[-2.27, 9.09]$ & $+7.50\;[3.50, 11.71]$ \\
 & CaM & $+2.80\;[0.00, 5.60]$ & $-1.14\;[-4.55, 2.27]$ & $+12.00\;[6.38, 17.50]$ \\
 & ChunkKV & $+0.40\;[-2.80, 3.80]$ & $-4.55\;[-15.91, 5.68]$ & $+5.50\;[1.50, 9.52]$ \\
\midrule
\multirow{6}{*}{\shortstack[l]{Qwen2\\7B-Instruct}} & BM25 & $+8.60\;[4.60, 12.60]$ & $+7.95\;[-4.55, 19.32]$ & $-3.50\;[-9.72, 2.44]$ \\
 & SnapKV & $+6.20\;[2.20, 10.00]$ & $+2.27\;[-4.55, 10.23]$ & $+14.50\;[5.85, 22.68]$ \\
 & PyramidKV & $+5.80\;[2.20, 9.40]$ & $+3.41\;[-2.27, 9.09]$ & $+13.50\;[5.11, 22.05]$ \\
 & H2O & $+4.60\;[1.60, 7.80]$ & $+4.55\;[-3.41, 12.50]$ & $+5.50\;[-0.47, 11.70]$ \\
 & CaM & $+5.00\;[1.80, 8.20]$ & $+1.14\;[-6.82, 9.09]$ & $+12.00\;[3.52, 19.31]$ \\
 & ChunkKV & $+9.40\;[5.60, 13.40]$ & $+1.14\;[-12.50, 12.50]$ & $+2.50\;[-5.26, 10.43]$ \\
\bottomrule
\end{tabular}
\caption{Paired score differences (\methodname{} minus Original), in percentage points, with 95\% cluster-bootstrap confidence intervals in brackets.}
\label{tab:paired_statistics}
\end{table*}

%% file: sections/appendix_position_evaluation.tex
\section{Alignment under Heterogeneous Cache Capacities}
\label{app:endpoint_alignment}
Sparse selectors can retain different numbers of states across layers. A shared question start therefore needs to coordinate several historical endpoints. We compare two compact, source-order mappings: RankSync assigns each retained sequence positions $0,\ldots,m_{\ell h}-1$, while \methodname{} assigns $M-m_{\ell h},\ldots,M-1$, with $M=\max_{\ell,h}m_{\ell h}$. Both start the question at $M$ and advance subsequent coordinates consecutively.

The mappings organize the history--question boundary differently. Under RankSync, the last retained key in a shorter cache is $M-m_{\ell h}+1$ positions from the first question token. Endpoint alignment makes this distance one in every layer. Internal retained-state spacing and source chronology remain the same.

\subsection{Controlled Comparison}
We use 200 LongMemEval-S questions, Llama-3.1-8B-Instruct and Qwen2-7B-Instruct, and BM25 and PyramidKV at 10\% history retention. Each comparison shares the selected historical states, values, masks, order, and actual budget. Original preserves source coordinates. RankSync and \methodname{} use the compact mappings above. Accuracy is scored by Qwen2.5-32B, and token-F1 provides a complementary answer measure. This 200-question comparison focuses on alignment at a fixed retention ratio. Its cohort is distinct from the full main-table evaluation.

\input{tables/e074_endpoint_alignment}

\paragraph{Capacity and continuation geometry.}
For BM25, retained lengths are uniform and $m_{\ell h}=M$, so both compact mappings reduce to $p_i'=i$. Their cache transformations and continuations coincide, and the identical RankSync and \methodname{} entries in Table~\ref{tab:endpoint_alignment} share the same computation.

PyramidKV produces unequal retained lengths across layers. Endpoint alignment increases judged accuracy over RankSync from 21.5\% to 22.5\% for Llama and from 10.5\% to 20.5\% for Qwen. Token-F1 increases by 1.37 and 1.14 percentage points, respectively. In these configurations, physical KV heads within a layer have equal retained counts. The comparison thus examines the continuation boundary across layers. The results illustrate how cache capacity and positional continuation interact: compacting each sequence and connecting its endpoint to the current input are separate design choices. This fixed-ratio comparison supports endpoint alignment for heterogeneous capacities in the evaluated configurations. It does not by itself explain performance changes across retention budgets.

%% file: tables/e074_endpoint_alignment.tex
\begin{table}[!htbp]
\centering
\small
\setlength{\tabcolsep}{3pt}
\begin{tabular}{llrrrr}
\toprule
\multicolumn{6}{c}{\textbf{(a) Accuracy (\%)}} \\
\midrule
Selector & Model & O & R & T & $\Delta$ \\
\midrule
BM25 & Llama & 24.0 & 38.5 & 38.5 & 0.0 \\
BM25 & Qwen & 21.0 & 36.0 & 36.0 & 0.0 \\
PyramidKV & Llama & 21.5 & 21.5 & \textbf{22.5} & +1.0 \\
PyramidKV & Qwen & 13.0 & 10.5 & \textbf{20.5} & +10.0 \\
\midrule
\multicolumn{6}{c}{\textbf{(b) Token-F1 (\%)}} \\
\midrule
Selector & Model & O & R & T & $\Delta$ \\
\midrule
BM25 & Llama & 9.31 & 11.68 & 11.68 & 0.00 \\
BM25 & Qwen & 5.89 & 8.93 & 8.93 & 0.00 \\
PyramidKV & Llama & 8.49 & 8.75 & \textbf{10.12} & +1.37 \\
PyramidKV & Qwen & 3.90 & 4.61 & \textbf{5.75} & +1.14 \\
\bottomrule
\end{tabular}
\caption{Coordinate alignment on fixed caches: 200 questions per configuration at 10\% history retention. O: Original, R: RankSync, T: \methodname{}. $\Delta$ denotes T$-$R in percentage points, calculated before rounding. Bold marks improvements over R.}
\label{tab:endpoint_alignment}
\end{table}

%% file: tables/e076_protocol_sensitivity.tex
\begin{table}[H]
\centering
\small
\setlength{\tabcolsep}{5pt}
\begin{tabular}{lrrr}
\toprule
Selector & \shortstack{$\Delta$\\Qwen} & \shortstack{$\Delta$\\gpt-oss} & \shortstack{gpt-oss\\O $\rightarrow$ T} \\
\midrule
BM25 & +8.60 & +10.20 & 28.00\,$\rightarrow$\,38.20 \\
SnapKV & +6.20 & +1.80 & 18.60\,$\rightarrow$\,20.40 \\
PyramidKV & +5.80 & +2.20 & 18.00\,$\rightarrow$\,20.20 \\
H2O & +4.60 & +4.00 & 11.20\,$\rightarrow$\,15.20 \\
CaM & +5.00 & +0.20 & 9.60\,$\rightarrow$\,9.80 \\
ChunkKV & +9.40 & +6.40 & 20.00\,$\rightarrow$\,26.40 \\
\bottomrule
\end{tabular}
\caption{Qwen2-7B-Instruct/LongMemEval-S, 10\% history retention, 500 pairs per selector. Accuracy (\%), $\Delta$: \methodname{}$-$Original (pp). O: Original, T: \methodname{}.}
\label{tab:protocol_sensitivity}
\end{table}